\documentclass[conference]{IEEEtran}
\IEEEoverridecommandlockouts
\usepackage{cite}
\usepackage{amsmath,amssymb,amsfonts}
\usepackage{algorithmic}

\usepackage{graphicx}
\usepackage{tabularx}
\usepackage{textcomp}
\usepackage{array}
\usepackage{makecell}

\def\BibTeX{{\rm B\kern-.05em{\sc i\kern-.025em b}\kern-.08em
    T\kern-.1667em\lower.7ex\hbox{E}\kern-.125emX}}

\begin{document}
\font\myfont=cmr12 at 20pt
\title{{\myfont Posture recognition using an RGB-D camera : exploring 3D body modeling and deep learning approaches}
\thanks{*This research was supported by the Canada Research Chair on Biomedical Data Mining  (950-231214).}
}

\author{\IEEEauthorblockN{Mohamed El Amine Elforaici\IEEEauthorrefmark{1},
Ismail Chaaraoui\IEEEauthorrefmark{1},
Wassim Bouachir\IEEEauthorrefmark{1},
Youssef Ouakrim\IEEEauthorrefmark{1} and
Neila Mezghani\IEEEauthorrefmark{1}}

\IEEEauthorblockA{\IEEEauthorrefmark{1}Laboratoire de recherche en Imagerie et Orthop\'edie (LIO), T\'ELUQ University, Montr\'eal, Canada}
}

\maketitle
\maketitle

\begin{abstract}
The emergence of RGB-D sensors offered new possibilities for addressing complex  artificial vision problems efficiently. Human posture recognition is among these computer vision problems, with a wide range of applications such as ambient assisted living and intelligent health care systems. In this context, our paper presents novel methods and ideas to design automatic posture recognition systems using an RGB-D camera. More specifically, we introduce two supervised methods to learn and recognize human postures using the main types of visual data provided by an RGB-D camera. The first method is based on convolutional features extracted from 2D images. Convolutional Neural Networks (CNNs) are trained to recognize human postures using transfer learning on RGB and depth images. Secondly, we propose to model the posture using the body joint configuration in the 3D space. Posture recognition is then performed through SVM classification of 3D skeleton-based features. To evaluate the proposed methods, we created a challenging posture recognition dataset with a considerable variability regarding the acquisition conditions. The experimental results demonstrated comparable performances and high precision for both methods in recognizing human postures, with a slight superiority for the CNN-based method when applied on depth images. Moreover, the two approaches demonstrated a high robustness to several perturbation factors, such as scale and orientation change. 
\end{abstract}


\section{Introduction}
Posture recognition is a central problem in computer vision, with a wide range of applications such as human-machine interaction, content-based video annotation, and especially intelligent video surveillance systems for ambient assisted living. During recent years, several research works tried to address the posture recognition problem following various machine vision methods and using different types of visual data. Traditional posture recognition systems used conventional cameras to build a body model from RGB images through background subtraction or geometric transformations \cite{1, 2, 3}. These methods present inherent limitations and lack the flexibility to handle challenging cases, such as dynamic backgrounds, camera motion, and silhouette deformation. Moreover, analyzing bi-dimensional images to construct a 3D body model, such as in \cite{3}, is often a source of computational complexity.

\par
The emergence of RGB-D cameras (e.g. Microsoft Kinect) allowed to exploit depth information in order to design less complex methods for modeling human postures. In this sense, recent works tried to extend 2D image features to the 3D domain for human body modeling. Most of these works relied on the Space Time Interest Points (STIPs) \cite{4}, an extension of the 2D interest points into the spatio-temporal domain.The STIP framework were implemented in different ways, especially using the Harris3D detector \cite{4} to identify interest points and the HOGHOF descriptor \cite{5} to describe the regions around the STIPs. 
   
\par
Instead of generalizing 2D image descriptors to the 3D space, a second line of works \cite{6, 7, 8} propose to use the skeleton detector \cite{9} to build high-level features characterizing the 3D configuration of the body. This representation was successfully applied for activity recognition as it conforms to the early Johansson's biological studies on how human understand actions \cite{10}.

\par
This paper aims to explore several acquisition possibilities offered by an RGB-D camera to design novel methods and ideas for automatic posture recognition. Our work proceeds along the two directions described above to propose and analyze two methods for posture recognition. Firstly, we propose a posture recognition method from 2D images. Instead of using hand-crafted features such as STIPs, we train Convolutional Neural Networks (CNNs) on RGB and depth images to recognize body postures. To the best of our knowledge, this is the first study proposing to use CNNs for static pose recognition. We note that deep neural networks were previously used for hand pose and gesture recognition \cite{11, 12}. Secondly, we propose a new method for body modeling using joint-based features as a high-level descriptor. The proposed method uses a single depth image to perform supervised classification based on the 3D skeleton spatial configuration. We finally provide an experimental evaluation of the proposed methods on a new posture RGB-D dataset created using the Microsoft Kinect v2 camera.

\par The rest of the paper is organized as follows: section \ref{sec:methods} presents the proposed methods, section \ref{sec:results} provides experimental results and discussions. Finally, section \ref{sec:conclusion} concludes the paper.

\section{The Proposed Methods}
\label{sec:methods}

\subsection{\textbf{Posture recognition using 2D images and Convolutional Neural Network}}

The proposed method is based on using 2D images representing human postures to adapt a pretrained CNN for our specific recognition problem. We use the AlexNet network \cite{13}, that has been pre-trained from the ImageNet database to classify images into 1000 object categories. Before using posture images to retrain the network, our dataset is first pre-processed by resizing the images according to the CNN input format and subtracting the background to extract the body silhouette. Silhouette segmentation is initially performed through body parsing on depth images \cite{9}. The segmentation result is then mapped to RGB images, since the infrared and RGB sensors are geometrically calibrated. 

\par
\begin{figure}
\centering
\includegraphics [width=8cm]{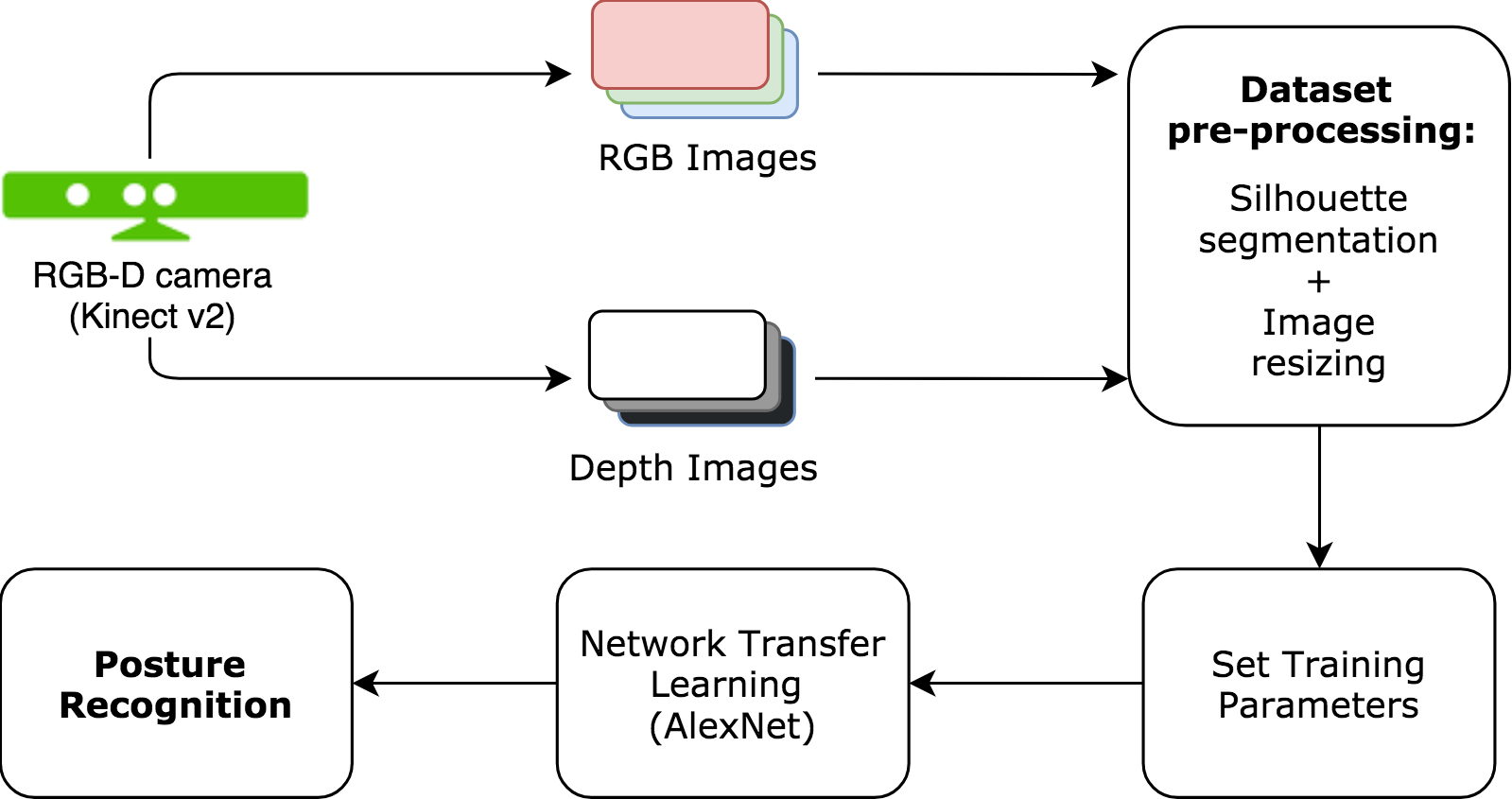}
\caption[fig1]
{\label{fig1} Posture recognition using 2D images and CNN}
\end{figure}

Once the pre-processing step is completed, we use the resulting images to retrain the CNN for posture classification. Prior to the network training procedure, we fine-tune the model in order to adapt the network to the new dataset. Since, training is performed using a relatively small image collection, the weights on initial layers are frozen. However, the last three layers were transferred to the new classification task by replacing them with a fully connected layer, a softmax layer, and a classification output layer \cite{13}. The classification output layer was set according to the five defined posture classes: standing, bending, sitting, walking, and crouching. The main steps of the proposed method are summarized in figure \ref{fig1}.  

\vspace{-0.5em}

\subsection{\textbf{Posture recognition using 3D joint-based features}}

\begin{figure}
\centering
\includegraphics [width=4cm]{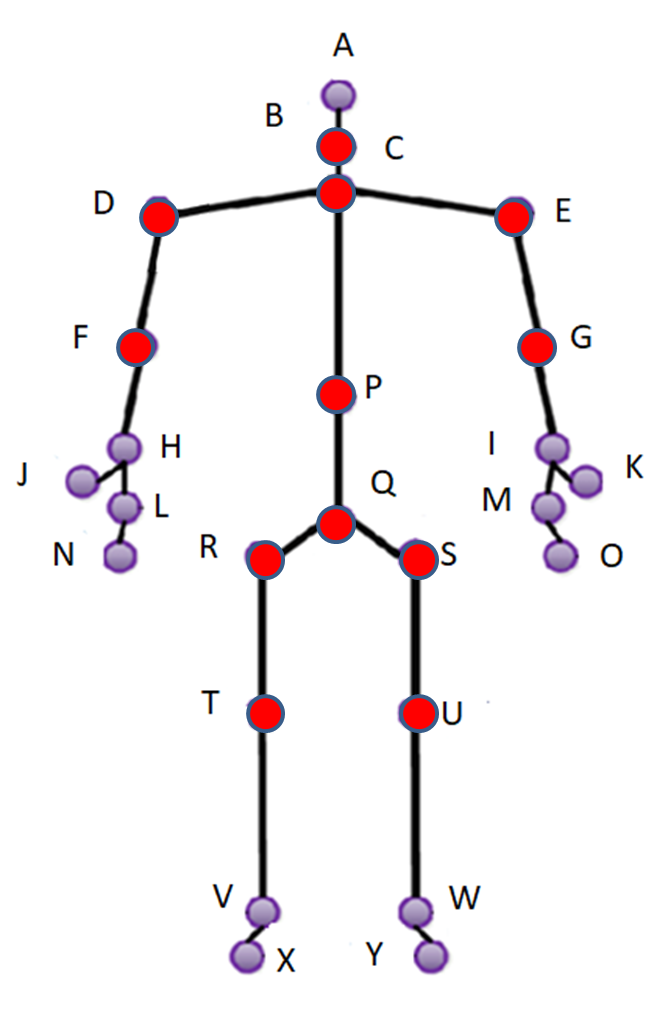}
\caption[fig2]
{\label{fig2} The 3D joints defining the skeleton.}
\end{figure}

The second method is based on classifying high-level features representing the 3D body configuration. Firstly, a 3D skeleton model is built using the skeleton detector of Shotton et al. \cite{9} for localizing different body parts and accurately model an articulated structure of connected segments. The obtained 3D skeleton formed by 25 joints represents the human posture as illustrated in the figure \ref{fig2}. Once joint positions are extracted, we propose to compute 2 types of features representing human body posture: the 3D pairwise distances between joints, and the geometrical angles of adjacent segments.

\par Given the body skeleton obtained from depth images, we calculate the relative 3D distances between pairs of joints. Such features were successfully used to model human body for action recognition \cite{6}.
To ensure scale invariance, the 3D distances are normalized with respect to the person's height. The normalization process is performed by considering the 3D distance between the spine shoulder joint and the spine middle joint. This distance was chosen based on experimental observations concluding that it is sufficiently stable with respect to several body deformations, in addition to its proportionality to the person's height.

\par
\begin{figure}
\centering
\includegraphics[width=9cm]{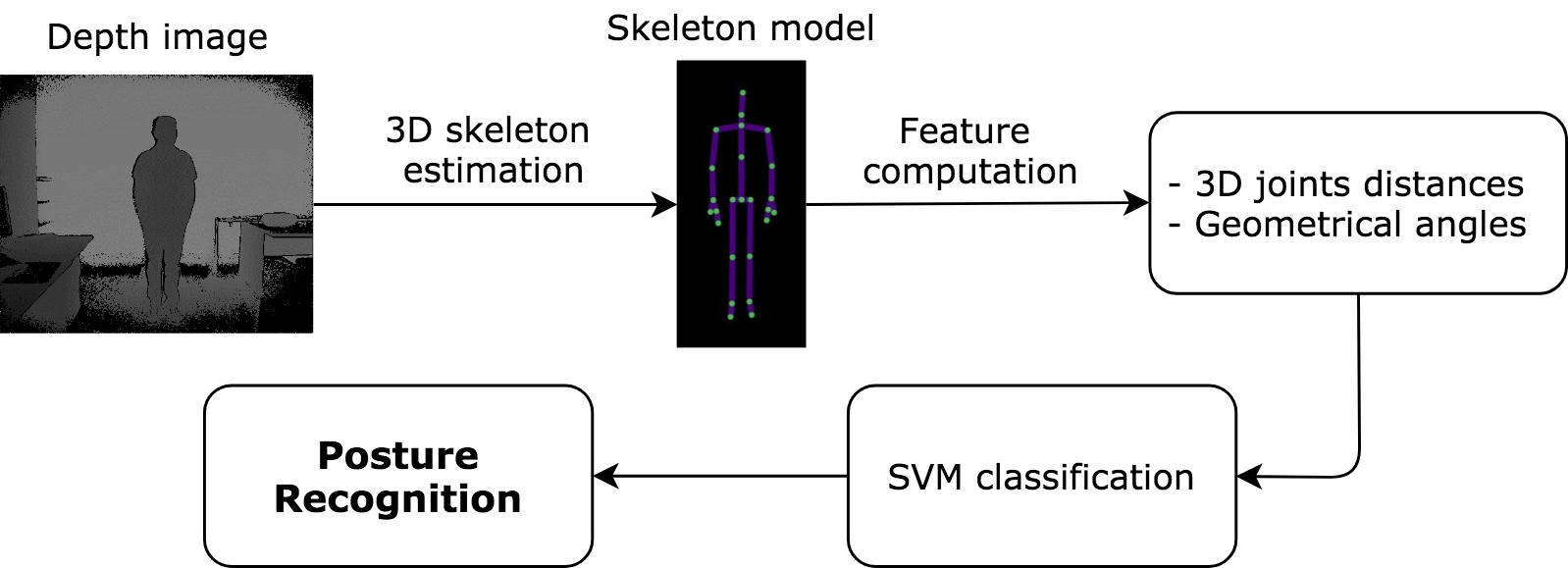}
\caption[fig1]
{\label{fig3} Posture recognition using 3D joint-based features}
\end{figure}

 In addition to 3D joint distances, our joint-based features include geometrical angles extracted from every possible combination of three joints on the skeleton model. Geometrical angles defined by adjacent segments are directly estimated from joint positions in the 3D space. Once the body pose is modeled, posture recognition is performed using SVM classification. Figure \ref{fig3} illustrates the main modules of the proposed system.

\section{Experiments}
\label{sec:results}
\subsection{\textbf{Data collection}}

\begin{figure}
\centering
\includegraphics [width=8.5cm]{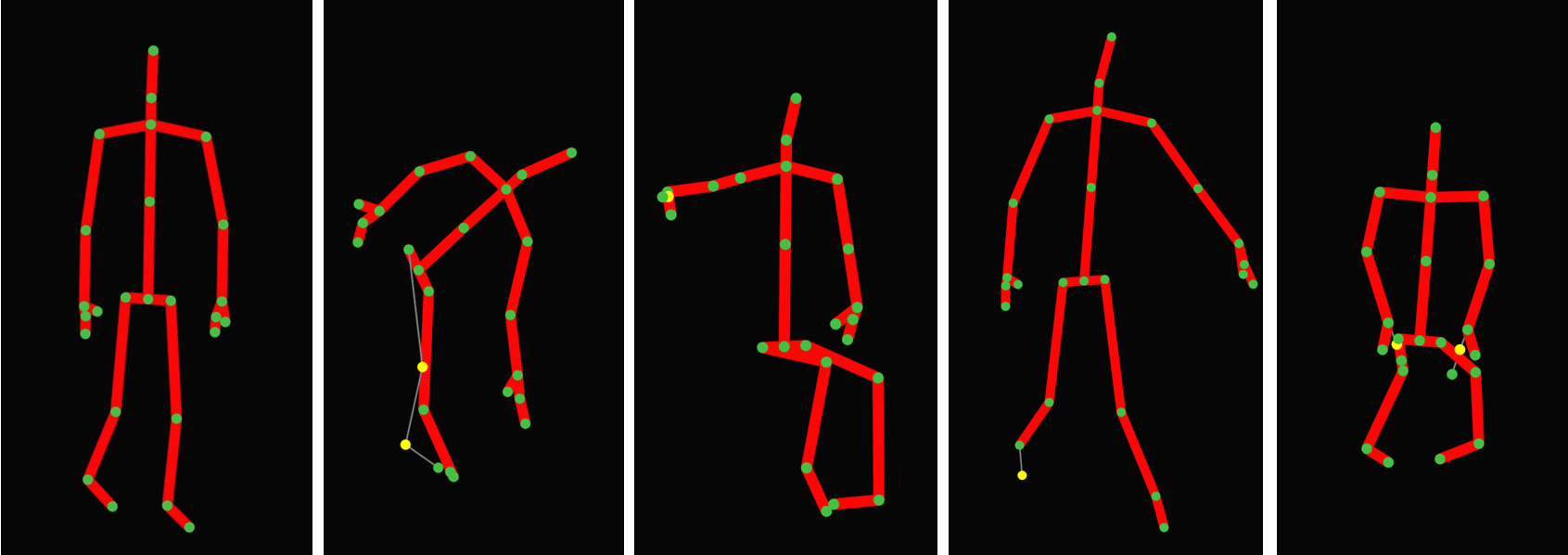}
\caption[fig4]
{\label{fig4} Examples of body skeletons. From left to right : standing, bending, sitting, walking, and crouching.}
\end{figure}

To evaluate the performance of the proposed methods, we created a new RGB-D dataset using the Microsoft Kinect v$2$ camera. The dataset includes the 3D joint coordinates in addition to RGB images and depth maps for each observation. To the best of our knowledge, this is the first posture recognition dataset created with the second version of the Kinect sensor (using Time of flight technology for depth sensing)\footnote{The dataset will be published online upon acceptance}.\\ In our experimental setup, the camera is mounted on a tripod placed in a corner of our laboratory. Figure \ref{fig4} illustrates five examples of body skeleton representing the five classes of postures: standing, bending, sitting, walking, and crouching. Each posture was performed by $13$ participants of different ages, gender, and morphological characteristics. In order to increase the variability of our dataset and evaluate our system in handling scale and orientation change, each posture is captured at four different orientations and four different distances with respect to the camera. The distance varies from one to four meters while the orientation angle varies between 0 and 360 degrees. Our dataset includes a total number of 1040 observations that we used for training and testing the proposed system. For both methods, the dataset was split into two subsets, where 80\%  was used for training the systems, while the remaining 20\% was exclusively used for testing.

All the experiments were implemented in Matlab and were executed on a machine with an Intel Core i7 2.93GHz processor and 8GB of RAM.\\
We designed our experiments in order to analyze the performance of our method according to several aspects.

\subsection{\textbf{Experimental results}}
\subsubsection{\textbf{Posture recognition using 2D images and CNN}}

\par The first experiment aimed to identify the optimal image type, among RGB and depth images, for CNN training and classification. Table \ref{tab1} presents the testing results for CNN models trained on segmented body silhouettes. The results show a slight superiority in accuracy for depth images.

\begin{table}
\caption {Validation and testing results of segmented RGB and depth images.}
\centering

\centering
\begin{tabular}{|c|c|}
\hline
\hline
{Image Type} & {Testing results}\\
\hline
\hline
\textit{RGB} & 93.3\%\\
\hline
\textit{Depth} & 95.7\%\\
\hline

\end{tabular}

\label{tab1}
\end{table} 

\par To further analyze the performance of the proposed methods, we used the confusion matrices. This evaluation tool provides a comparison of reference postures to the predicted postures. The rows of the matrices represent the true classes, while the columns represent the predicted classes. As shown in tables \ref{tab2} and \ref{tab3}, high recognition results were achieved for almost all the classes. We observed a 100\% accuracy for the bending, sitting and crouching classes. On the other hand, there was substantial confusion between the standing and walking classes with an accuracy rate of respectively $78.6\%$ and $88.1\%$ for RGB images, and $92.9\%$ and $85.7\%$ for depth images. This confusion can be explained by the visual similarity in the silhouette appearance between the standing and walking postures.

\begin{table}
\caption {Posture recognition matrix: CNN classification of RGB images.}
\centering
\centering
\begin{tabular}{|c|c|c|c|c|c|}
\hline
\hline
Postures & \textit{Standing} & \textit{Bending} & \textit{Sitting} & \textit{Walking} & \textit{crouching}\\
\hline
\hline
\textit{Standing} & \textbf{78.6\%} &0 \% &0\% &21.4\% &0\% \\
\hline
\textit{Bending} &0\% &\textbf{100\%} &0\% &0\% &0\% \\
\hline
\textit{Sitting} &0\% &0 \% &\textbf{100\%} &0\% &0\% \\
\hline
\textit{Walking}& 11.9\% &0 \% &0\% &\textbf{88.1\%} &0\% \\
\hline
\textit{crouching}& 0\% &0 \% &0\% &0\% &\textbf{100\%} \\
\hline
\end{tabular}

\label{tab2}
\end{table}

\begin{table}
\caption {Posture recognition matrix: CNN classification of depth images.}
\centering
\centering
\begin{tabular}{|c|c|c|c|c|c|}
\hline
\hline
Postures & \textit{Standing} & \textit{Bending} & \textit{Sitting} & \textit{Walking} & \textit{crouching}\\
\hline
\hline
\textit{Standing} & \textbf{92.9\%} &0 \% &0\% &7.1\% &0\% \\
\hline
\textit{Bending} &0\% &\textbf{100\%} &0\% &0\% &0\% \\
\hline
\textit{Sitting} &0\% &0 \% &\textbf{100\%} &0\% &0\% \\
\hline
\textit{Walking}& 14.3\% &0 \% &0\% &\textbf{85.7\%} &0\% \\
\hline
\textit{crouching}& 0\% &0 \% &0\% &0\% &\textbf{100\%} \\
\hline
\end{tabular}

\label{tab3}
\end{table}

\vspace{0.5cm}

\subsubsection{\textbf{Posture recognition using 3D joint-based features}}

\par Our first experiment set aimed to identify the optimal choices regarding the two main system components: 1) the feature set and, 2) the classification method. We thus investigated several choices entailing five classifiers (linear discriminant, quadratic discriminant, linear SVM, quadratic SVM, and cubic SVM), and three feature sets (geometrical angles, pairwise joint distances, and the combination of the 2 types of features). Table \ref{tab4} summarizes accuracy results for the tested classifiers applied to different feature sets. We can remark that the best recognition rates of linear discriminant and Fine KNN classifiers were obtained by using the 3D joints distances as features. However, SVM classifiers clearly outperformed both methods regardless the feature set. This investigation shows that the Quadratic SVM applied to the entire feature set achieves the best recognition rate ($93.1\%$). We therefore base our recognition system on 1) Quadratic SVM classification, and 2) a combination of the two types of features.

\begin{table}
\caption {Performance results of several classifiers using different types of features.}
\centering
\centering
\begin{tabular}{|c|c|c|c|}
\hline
\hline
\thead{Classifier} &  \thead{Geometrical \\ angles} &  \thead{3D joints \\ distances} & \thead{Combined \\ features} \\
\hline
\hline
\textit{Linear Discriminant} & 83.3\%  & 87.4\% & 83.2\% \\
\hline
\textit{Fine KNN} & 84.5\% & 87.7\% & 86.2\% \\
\hline
\textit{Linear SVM} & 89.2\% & 89.8\% & 91.2\% \\
\hline
\textit{Quadratic SVM} & 92.3\% & 92.3\% & \textbf{93.1\%} \\
\hline
\textit{Cubic SVM} & 91.7\% & 92.0 \% & 92.5\% \\
\hline
\end{tabular}
\label{tab4}

\end{table}

 By analyzing the corresponding confusion matrix, it can be seen that the diagonal elements show high classification rates for all the classes. In table \ref{tab5}, the classification accuracy is higher than $90\%$ for almost all the classes, except for the walking class ($89.9\%$). The highest classification accuracy was observed for the sitting class ($97.6\%$), since this posture is characterized by distinctive joint-based features. As with the first method, a confusion was also noted between the two classes walking and standing. However, the percentages of confusion are lower than those of the first method, which suggests that posture recognition using 3D joint-based features is more effective in distinguishing between the walking and standing postures.

\begin{table}
\caption {Posture recognition matrix for quadratic SVM classification using combined features.}
\centering
\centering
\begin{tabular}{|c|c|c|c|c|c|}
\hline
\hline
Postures & \textit{Standing} & \textit{Bending} & \textit{Sitting} & \textit{Walking} & \textit{crouching}\\
\hline
\hline
\textit{Standing} & \textbf{92.3\%} &0 \% &0\% &7.7\% &0\% \\
\hline
\textit{Bending} & 0.5\% &\textbf{91.8\%} &1\% &4.8\% &1.9\% \\
\hline
\textit{Sitting} & 0.5\% &0\% &\textbf{97.6\%} &0\% &1.9\% \\
\hline
\textit{Walking}& 8.6\% &0.5\% &1\% &\textbf{89.9\%} &0\% \\
\hline
\textit{crouching}& 0\% &1.0 \% &5.3\% &0\% &\textbf{93.7\%} \\
\hline
\end{tabular}

\label{tab5}
\end{table}

\subsection{\textbf{Discussion}}
\label{sec:discussion}

Due to their high accuracy and low computational complexity, both approaches can be applied for implementing practical applications, while meeting real-time requirements. However, several design choices depend on specific use cases and environmental constraints. For example, depth images lose their mapping quality at distances over three meters from the RGB-D camera. They are therefore inappropriate for open environments. We also observed a degradation in the silhouette segmentation quality on the two types of images in this situation. This can be explained by the fact that the segmentation method is highly dependent on depth maps.

From another point of view, posture recognition performed using depth maps and 3D skeleton model remain efficient regardless of lighting conditions, as the corresponding features only dependent on the infrared sensors of the RGB-D camera. This makes the system more robust and suitable for indoor environments where lighting condition is not stable. On the other hand, this type of environment represents a major limitation for RGB-based methods that are generally sensitive to light changes.

\vspace{2mm}
\section{Conclusion}
\label{sec:conclusion}

We presented novel solutions for automatic posture recognition using an RGB-D camera. We mainly designed two methods by exploiting different types of visual data provided by an RGB-D camera. The first method is based on CNN classification of 2D images, while the second uses 3D body modeling to compute and classify high level features. The proposed methods were validated on a challenging posture recognition dataset released in our laboratory. The experimental results showed comparable performances and high precision for both methods, with a slight superiority for the CNN-based method when applied on depth images. Moreover both methods demonstrated a significant invariance to important perturbation factors, including scale and orientation changes. 

This study offers several choices for implementing practical posture recognition systems, depending on the use cases and application constraints. As an example, methods using depth images are suitable for indoor environments with limited space. In such an environment, pose recognition can be efficiently achieved regardless the illumination condition.

\end{document}